\documentclass[11pt,oneside,letter]{article}
\usepackage[top=1 in, bottom=1 in, left=0.85 in, right=0.85 in]{geometry}

\usepackage{amsmath,amsfonts,bm}

\def\eqref#1{equation~\ref{#1}}

\def\1{\bm{1}}

\DeclareMathAlphabet{\mathsfit}{\encodingdefault}{\sfdefault}{m}{sl}
\SetMathAlphabet{\mathsfit}{bold}{\encodingdefault}{\sfdefault}{bx}{n}

\usepackage{float}
\usepackage{times}
\usepackage{mathtools}
\usepackage[round]{natbib}
\usepackage{makecell}
\usepackage{amsmath,amssymb,graphicx,url}
\usepackage{thmtools,thm-restate,wrapfig,enumitem,mathabx}
\usepackage{booktabs}
\usepackage{color}
\usepackage{subfigure}
\usepackage{colortbl}
\usepackage{multirow}

\usepackage[utf8]{inputenc} 
\usepackage[T1]{fontenc}    
\usepackage{hyperref}       
\usepackage{url}            
\usepackage{booktabs}       
\usepackage{amsfonts}       
\usepackage{nicefrac}       
\usepackage{microtype}      
\usepackage{xcolor}         

\usepackage{algorithm}
\usepackage[noend]{algorithmic}
\usepackage{titletoc}
\usepackage{bbm}

\newtheorem{assumption}{Assumption}

\newtheorem{proof}{Proof}

\begin{document}

\title{
Actor-Predictor Architecture for Reinforcement Learning with Formal Specification Objectives
}
\date{}

\author{
	Josef Brozovich \\
	University of Michigan, Ann Arbor \\  \texttt{broz@umich.edu} \\ 
	\and
    Qining Zhang \\
	University of Michigan, Ann Arbor \\  \texttt{qiningz@umich.edu} \\ 
	\and
	Lei Ying \\
	University of Michigan, Ann Arbor \\  \texttt{leiying@umich.edu}\\
	\\
}

\maketitle

\begin{abstract}

    One fundamental challenge in reinforcement learning (RL) is enforcing temporal safety constraints, such as satisfying formal specifications. Existing methods often transform logic formulas into automata for state augmentation and reward shaping. However, this approach assumes that automaton construction is both feasible and tractable, which limits its applicability to small specifications and linear temporal logic (LTL).
    This paper proposes Actor-Predictor, a framework that learns a surrogate representation of logic progress without explicitly constructing an automaton. The predictor continuously estimates whether a partial trajectory will satisfy the formal specification in a later step, and its learned features are provided as auxiliary inputs to the actor, which is then optimized using GRPO and RWR. We show theoretically that, under sufficient predictive accuracy, the learned representation is equivalent to the automaton state.
    Experiments on grid world and robotic tasks in Isaac Lab demonstrate that Actor-Predictor outperforms automaton-based baselines on LTL-constrained tasks. It also performs well on non-regular formal specifications, such as $(a^n b^n)$, for which finite-automaton-based approaches are not applicable.
\end{abstract}

\section{Introduction}
Formal specification is a framework for describing complex, time-dependent behaviors, such as safety constraints. 
Reinforcement learning (RL) with formal specifications is a data-driven approach to train an agent to learn policies that optimize objectives described by temporal logic formulas or non-regular specifications, which faces two fundamental challenges. 
First, formal specifications introduce long-range temporal dependencies, meaning that the optimal policy is generally non-Markovian, which may depend on the history of observations and actions rather than solely on the current environment state. 
Second, formal specification objectives yield binary, trajectory-level feedback, causing extreme reward sparsity that results in challenging cold-starts. 

The standard approach to overcome this challenge is through automata theory and reward shaping~\citep{shah2024ltl, manganaris2025automaton}, which converts a temporal logic (TL) formula into a finite-state automaton and then optimizes the TL objective over the product state space including both the automaton state and the environment state. It often further modifies the reward function to provide dense feedback throughout the trajectory. 
However, it has been shown that translating an LTL formula into a limit-deterministic B\"uchi automaton (LDBA) can be computationally infeasible, where in the worst case, the complexity is {double-exponential} in the number of Atomic Propositions (APs)~\citep{sickert2016limit}. 
Furthermore, any language requiring an arbitrary counter is non-regular and, therefore, cannot be represented by a finite automaton. 

This paper takes a different approach. Our algorithm learns a surrogate representation of the automaton state using the features from a {\em predictor} that predicts the future satisfaction probability of a given partial trajectory. 
The intuition behind this approach is that an agent's progress through an automaton uniquely maps to a probability of reaching a future accepting state. 
The predictor network can be trained from a set of collected trajectories with binary satisfactory labels. The feature of the predictor, when given a partial trajectory, can represent the automaton state, thus avoiding the construction of a finite-state automaton.   
This leads to our proposed Actor-Predictor approach.

\subsection{Main contributions}
The key contributions of this paper include: 
\begin{itemize}[noitemsep, topsep=2pt, parsep=0pt, partopsep=0pt]
    \item \textbf{Theoretical Guarantees:} For deterministic automata with accepting states, we prove that: (i) a well-trained predictor's hidden representation sufficiently captures the automaton state (Lemma \ref{lem:predcitor}); and (ii) policy optimization over the surrogate product state yields an optimal policy for the formal specification objective (Theorem \ref{thm:policy}).
    \item \textbf{Algorithm \& Evaluation:} We implement Actor-Predictor using a Mamba predictor~\citep{GuDao_24} and a GRPO-optimized actor~\citep{guo2025deepseek} with exponential reward weighting. Across quadruped navigation (LTL) and grid-world (non-regular) tasks, experiments show: (1) Actor-Predictor outperforms baselines like CyclER~\citep{shah2024ltl}, (2) Mamba predictors outperform Transformer and LSTM baselines~\citep{vaswani2017attention, graves2012long}; (3) learned representations exhibit transferability and subformula compositionality; and (4) predictor features empirically suffice to reconstruct the automaton state.
\end{itemize}

\section{Related Work}

Combining formal specification with reinforcement learning to provide a formal framework for safety and task satisfaction is an active topic in RL~\citep{li2017reinforcement, bozkurt2020control, hasanbeig2020deep, hasanbeig2019reinforcement, li2018policy, cai2021modular, chou2022learning, wang2021reinforcement}. 
Temporal logic enables specifications over trajectories as a time-ordered sequence of steps. 
The standard approach to address RL in formal specification is to convert the Linear Temporal Logic (LTL) formula into a limit-deterministic B\"uchi automaton (LDBA). 
A product Markov decision process (PMDP) is then constructed by combining the system's state space and the B\"uchi Automaton, yielding the product state space $S_{\phi}=S \times  \mathcal{B}$.  
This transformation renders the task Markovian, enabling stationary policies over the augmented state space.

Using the LDBA, methods such as Automaton Constrained Q-Learning (ACQL), Cycle Experience Replay (CyclER), and LTL-guided Counterfactual Experience Replay (LCER) leverage progress within the automaton to automatically densify the reward signal~\citep{shah2024ltl, manganaris2025automaton, voloshin2023eventual}.
Although these methods improve learning efficiency, they fundamentally rely on access to the explicit structure and observability of the automaton. 
This restricts the expressivity of formal logic to regular languages. 
Consequently, tasks requiring arbitrary counting form non-regular languages and cannot be solved using these approaches.

Our approach is inspired by predictive state representations (PSRs) for POMDPs~\citep{LitSut_01}. In PSR, the latent state is represented by predictions of a collection of future tests, where each test corresponds to a sequence of future observations. 
While PSRs require carefully handcrafted tests, our predictor directly estimates the eventual satisfaction of the given specification.

\section{Problem Formulation}

\subsection{RL Problem}

This paper considers an RL problem defined by a tuple $\mathcal{M}=(\mathcal{S}, \mathcal{A}, \mathbb{P}, K, r_{\mathcal{L}})$, where $\mathcal{S}$ is the state space, $\mathcal{A}$ is the action space, $\mathbb{P}:\mathcal{S} \times \mathcal{A} \mapsto \mathcal{S}$ is the time-homogeneous transition kernel, $K$ is the time horizon, and $r_{\mathcal{L}}$ is the reward function defined by a formal specification $\mathcal{L}$ for trajectories, e.g., the accepting language $\mathcal{L}_{\varphi}$ of a linear TL (LTL) $\varphi$ or a non-regular formal specification. 
For example, the reward $r_{\mathcal{L}}(\tau)$ could be a simple binary feedback on whether the trajectory $\tau$ satisfies the formal specification, or the number of times the formal specification has been repeatedly satisfied, or the counter $n$ of the $\mathcal{L} = \{a^nb^n\}$ task.
Different from classic RL, our formal specification reward $r_{\mathcal{L}}$ is usually a function of the whole trajectory and cannot be decomposed into ``local'' rewards that depend on the current state-action pair. 
The objective of the agent is to find a policy $\pi^*$ that maximizes the expected cumulative trajectory reward, i.e., $\pi^* = \arg\max_{\pi} \mathbb{E}_{\tau \sim \pi}[r_{\mathcal{L}}(\tau)]$.

\subsection{Temporal Logic for Trajectories}

Consider a temporal logic formula $\varphi$, whose language $\mathcal{L}_\varphi$ or the set of trajectories that satisfy $\varphi$ defines a formal specification. 
A formal logic consists of atomic propositions (APs) and operators. Given a trajectory $\tau$ that consists of state-action pairs $\{(s_k, a_k)\}_{k=1}^K$, an AP is a Boolean-valued predicate that can be evaluated at each time step, i.e., $p:\mathcal{S}\times \mathcal{A}\mapsto \{\mathsf{True}, \mathsf{False}\}$.
Typical operators in TL that verify the characteristics of trajectories include: Finally $\mathbf{F}$, globally $\mathbf{G}$, next $\mathbf{X}$, until $\mathbf{U}$, and negation $\neg$. 
A temporal logic (formula) $\varphi$ is constructed from APs and logic operators. 
For example, let $A$, $B$, $C$, and $D$ be APs, a temporal logic formula is:
\begin{align}\label{eq:tl-example}
    \varphi=\mathbf{F}(A \land \mathbf{X}(\mathbf{F}(B \land \mathbf{X}(\mathbf{F}(C)))))\land \mathbf{G}(\neg D),
\end{align}
which specifies that an agent must loop around four states described by $A$, $B$, and $C$ in sequence without reaching a dangerous state described by $D$.

\begin{figure*}[t]
    \vspace{-10pt}
    \centering

    \begin{minipage}[t]{0.4\textwidth}
        \centering
        \includegraphics[width=\linewidth]{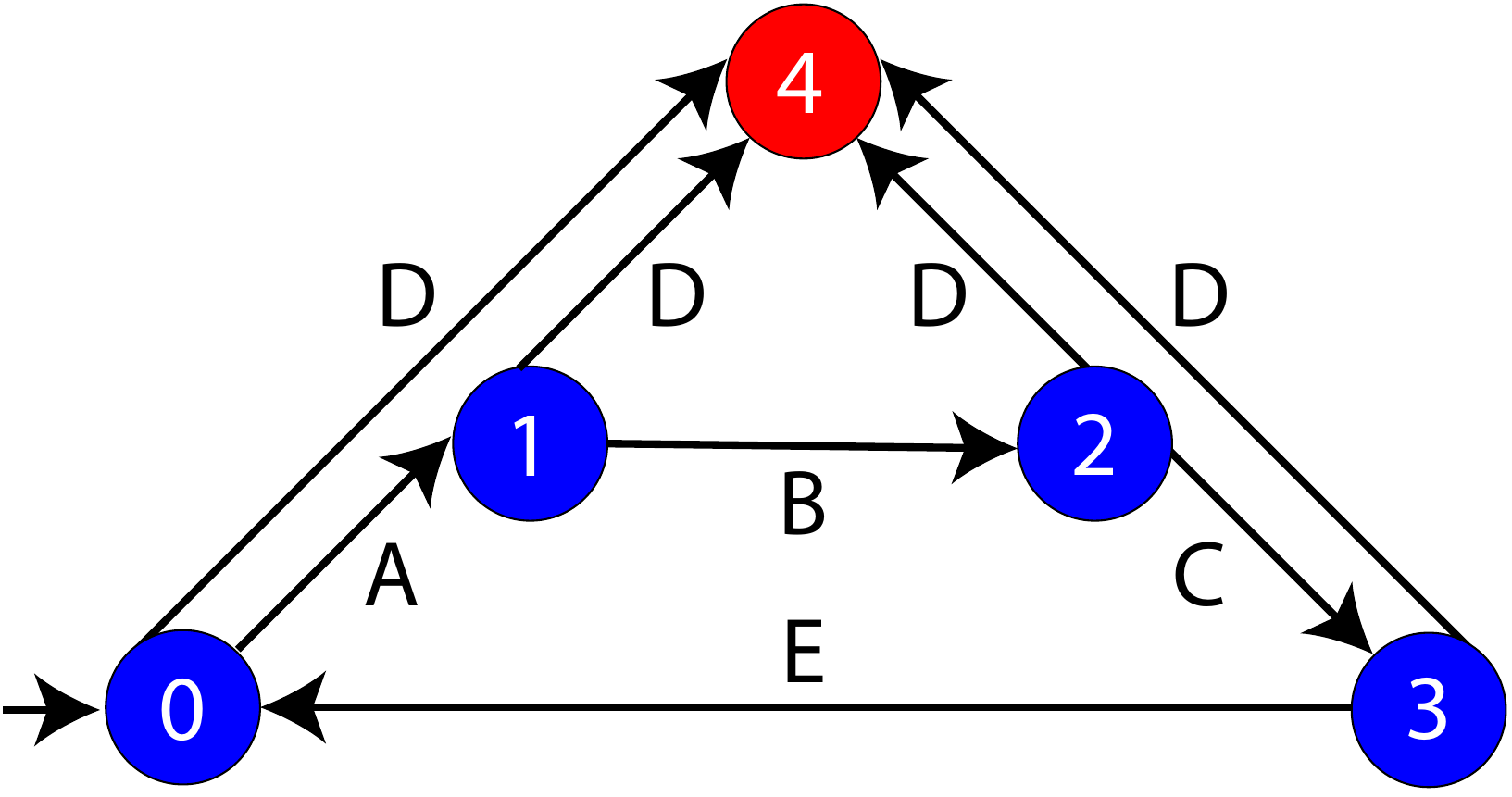}
        \vspace{-2mm}

        {\small (a)}
    \end{minipage}
    \hfill
    \begin{minipage}[t]{0.48\textwidth}
        \centering
        \includegraphics[width=\linewidth]{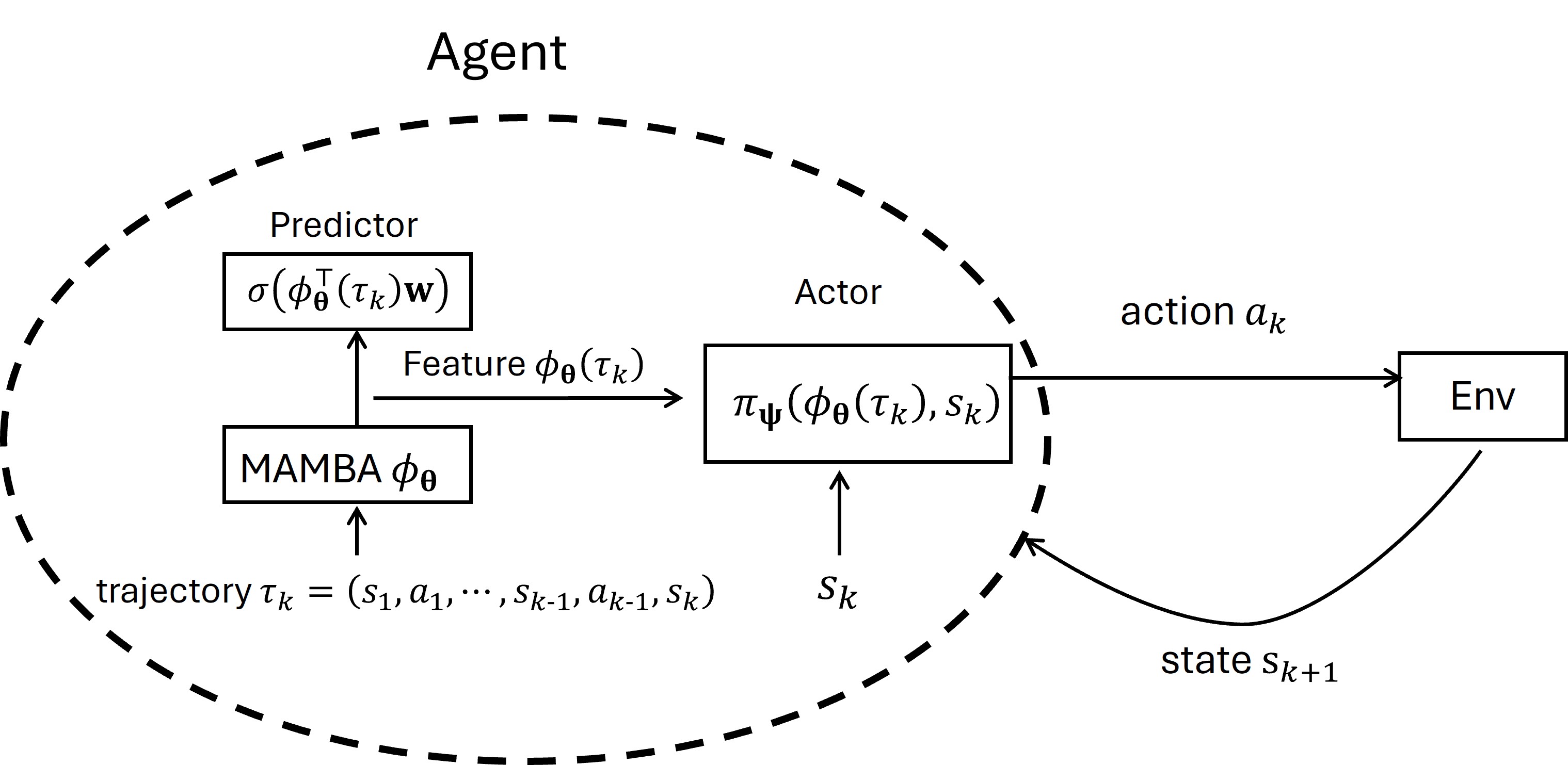}
        \vspace{-2mm}

        {\small (b)}
    \end{minipage}

    \caption{
    (a) Automaton corresponding to~\eqref{eq:tl-example}:
    reward is obtained upon completing
    $3 \to 0$; transition E denotes a reset.
    (b) Actor-predictor architecture with predictor
    $\phi_{\theta}$ and actor $\pi_{\psi}$.
    The predictor estimates satisfaction probability and passes
    features $\boldsymbol{\phi}_k$ to the actor.
    }
    \label{fig:framework-overview}
    \vspace{-10pt}
\end{figure*}

An LDBA~\citep{sickert2016limit} translates a TL formula into a state machine called {\bf automaton}. Given a TL formula $\varphi$ with APs and operators, an LDBA can be defined as the tuple $\mathbb{B} = (\mathcal{B}, \Sigma, T^{\mathbb{B}}, \mathcal{B}^{*}, b_{0})$, where $\mathcal{B}$ is the set of automaton states, $\Sigma$ is the power set of all APs in $\mathcal{L}$, $T^{\mathbb{B}}: \mathcal{B} \times (\Sigma \cup \{E\}) \rightarrow \mathcal{B}$ is the transition function, where $E$ is a dummy transition representing restart, $\mathcal{B}^{*}$ is the set of accepting states that define the acceptance condition of the LDBA, and $b_{0} \in \mathcal{B}$ is the initial state. An automaton example of the TL formula in~\eqref{eq:tl-example} is shown in Fig.~\hyperref[fig:framework-overview]{\ref*{fig:framework-overview}(a)}. 
By converting the LTL specification into an LDBA, the automaton state space $\mathcal{B}$ can be integrated with the RL state $\mathcal{S}$ to construct a product MDP ${\cal S}_\otimes={\cal S}\times {\cal B}$. Then, solving a TL task can be formulated as solving an MDP over the product space with state-action reward function $r:{\cal S}_\otimes\times{\cal A}\rightarrow \mathbb R^+$.
It has been shown that translating an LTL formula into an LDBA can be computationally infeasible, where in the worst case, the complexity is \textbf{double-exponential} in the number of APs~\citep{sickert2016limit}. 
In this paper, we take a different route and present an actor-predictor approach, inspired by {\em predictive state representation}~\citep{LitSut_01}, to mitigate these issues. In the rest of this paper, we assume the automaton is unknown but make the following assumptions:
\begin{assumption}[Deterministic Automaton with Accepting States]
The formal specification task $\mathcal{L}$ can be reduced to an unknown {\em deterministic} automaton, whose state is unobserved. 
The formal specification is satisfied when the state of the automaton is in one of the accepting states $\cal B^*$. In other words, a trajectory $\tau$ satisfies the formal specification if $b_k\in{\cal B}^*$ at some step $k.$. \label{assu:automaton}
\end{assumption}

\section{An Actor-Predictor Approach}

We name our architecture {\em Actor-Predictor}, which consists of a neural network predictor and an actor as shown in Fig.~\hyperref[fig:framework-overview]{\ref*{fig:framework-overview}(a)}. 
At step $k$, the predictor takes the entire history $\tau_k=(s_1, a_1, s_2, a_2, \cdots, s_k)$ as input and estimates the probability that, under the current policy, the formal specification $\varphi$ will be satisfied in a later step (after step $k$). Note that the predictor predicts {\em future} satisfactions, not past ones.
Specifically, let the feature vector output (second-to-last layer) of the predictor neural network be $\phi_{\bm \theta}(\tau_k)$, where $\boldsymbol{\theta}$ is the parameter, and let $\sigma(\cdot)$ be the logistic function. The predicted probability of satisfaction is 
$
    \sigma\left( \phi_{\bm \theta}(\tau_k)^\top \mathbf{w} \right)
$
where $\mathbf{w}$ are the linear weights of the last layer. 
For the actor, denoted as a probability distribution $\pi_{\bm \psi}$ where ${\bm \psi}$ is the actor parameter, it takes the feature $\phi_{\bm \theta}(\tau_k)$ of the predictor and the state $s_k$ as input. Then, it outputs an action sampled from $\pi_{\bm \psi}(\phi_{\bm \theta}(\tau_k), s_k)$. The predictor is trained with a binary cross-entropy loss, and the actor is trained using GRPO with reward-weighted regression (RWR). The full algorithm is summarized in Alg.~\ref{alg:actor_predictor}. 

Intuitively, to accurately predict the eventual satisfaction probability of the partial trajectory, the predictor feature should expressively represent useful formal specification features such as the automaton state. Integrating these features into the RL state as the input of the actor should reduce the history-dependent formal specification based decision-making problem to a stationary MDP problem, where the optimal policy is Markovian. We will prove this property under mild conditions. 

\begin{algorithm}[t]
\caption{Actor-Predictor Training}
\label{alg:actor_predictor}
\begin{algorithmic}[1]

\STATE \textbf{Input:} Initialize predictor parameters
$\boldsymbol{\theta}_0$, actor parameters
$\boldsymbol{\psi}_0$, and exponential weighting of $\beta$

\FOR{$t = 0,1,2,\dots$}

    \STATE Roll out the current policy as in
    Alg.~\ref{alg:actor_predictor_rollout} with predictor
    $\phi_{\boldsymbol{\theta}_t}$ and actor
    $\pi_{\boldsymbol{\psi}_t}$ to collect trajectories
    $\{\tau^{(n)}\}_{n=1}^N$ with rewards
    $r_{\mathcal{L}}(\tau^{(n)})$ and satisfaction labels
    $y_n$.

    \STATE Calculate raw advantages
    $\tilde{A}(\tau^{(n)})$ for all trajectories using inverse
    temperature $\beta$, and normalize them to obtain
    $A(\tau^{(n)})$ following~\eqref{eq:rwr}
    and~\eqref{eq:advantage-normalization}.

    \STATE Train the actor $\pi_{\boldsymbol{\psi}_t}$ with GRPO
    using loss~\eqref{eq:grpo}:
    $\boldsymbol{\psi}_{t+1}
    = \arg\min_{\boldsymbol{\psi}}
    \mathcal{L}_{\mathrm{GRPO}}(\boldsymbol{\psi})$.

    \STATE Train the predictor by minimizing the BCE loss:
    $\boldsymbol{\theta}_{t+1}
    = \arg\min_{\boldsymbol{\theta}}
    \mathcal{L}_{\mathrm{BCE}}(\boldsymbol{\theta})$.

\ENDFOR

\end{algorithmic}
\end{algorithm}

\subsection{Predictor with Mamba SSM}
Our predictor $\phi_{\bm{\theta}}$ estimates the probability of satisfaction given a partial trajectory $\tau_k=(s_1,a_1,\allowbreak s_2,a_2,\allowbreak \ldots,\allowbreak s_k)$ under the current policy.
We choose Mamba as our neural network architecture for the following reasons:
First, since the input is a time series, the predictor needs to have memory to capture the long-range dependency. 
Second, Mamba can dynamically change its internal state when an atomic proposition is satisfied, effectively mimicking the state transition of an automaton without requiring an explicit graph. 
Third, Mamba's computational complexity scales linearly with sequence length during training and remains constant per step during inference.
Unlike the discrete states of a standard automaton, Mamba does not restrict its hidden state to a binary representation, granting a significantly greater expressive power.

Given a set of partial trajectories $\{\tau_{k_n}^{(n)}\}_{n=1}^N$ where each trajectory consists of several consecutive state-action pairs from the initial state, and the binary labels $\{y_n\}_{n=1}^N$, where $y_n$ indicates whether  $\tau_{k_n}$ under the current policy would satisfy the formal specification $\mathcal{L}$ in a future step, (i.e., $b^{(n)}_{m} \in {\cal B}^*$ for some $K\geq m\geq k_n.$).  We train the Mamba predictor using the BCE loss as follows: 

\begin{equation}
    \mathcal{L}_{\text{BCE}}(\boldsymbol{\theta})=-\frac{1}{N}\sum_n \left[ y_n \log \sigma \left({\bm \phi}_{\bm \theta}\left(\tau_{k_{n}}^{(n)} \right)^\top {\textbf w}\right)+\right.\nonumber\left.(1-y_{n})\log \left(1-\sigma \left({\bm \phi}_{\bm \theta}\left(\tau_{k_{n}}^{(n)}\right)^\top{\textbf w}\right)\right)\right].\label{loss-bce}
\end{equation}

To obtain the partial trajectories, we could first roll out a complete trajectory with Alg.~\ref{alg:actor_predictor_rollout}, randomly sample some time steps $k_n$, and then verify its future satisfaction with formal specification.  

\textbf{Multi-predictor and subformulas.}
When a temporal logic specification is long, we can break it into subformulas and use a multi-Predictor Actor approach.
For each subformula on a partial trajectory $\tau_k$, we use a separate feature extractor $\phi_{\boldsymbol{\theta}_i}(\tau_k)$ to extract the feature of each subformula. Then, we independently verify whether each subformula is satisfied for the completion of the trajectory under the current policy. Similarly, we train all predictors using the BCE loss in~\eqref{loss-bce}.

\subsection{Training MLP Actor with GRPO and RWR}

\begin{algorithm}
\caption{Actor-Predictor Policy Roll-out}
\label{alg:actor_predictor_rollout}
\begin{algorithmic}[1]

\REQUIRE Predictor feature extractor $\phi_{\boldsymbol{\theta}}$, actor $\pi_{\boldsymbol{\psi}}$

\STATE Sample initial state $s_1$

\FOR{$k = 1,2,\ldots,K$}

    \STATE Extract predictor feature from history trajectory
    $\tau_k = \{s_1,a_1,\ldots,s_k\}$ to obtain
    $\boldsymbol{\phi}_k = \phi_{\boldsymbol{\theta}}(s_k)$

    \STATE Take action $a_k$ sampled from actor
    $\pi_{\boldsymbol{\psi}}(\cdot \mid s_k,\boldsymbol{\phi}_k)$
    and observe next state $s_{k+1}$

\ENDFOR

\STATE Let $\tau$ be the full trajectory generated; randomly sample $k$ to get partial trajectory $\tau_{k}$ and evaluate satisfaction in the future by checking the complete trajectory $\tau$ to get a corresponding label $y.$

\end{algorithmic}
\end{algorithm}

The actor $\pi_{\boldsymbol{\psi}}$ operates on an augmented state $\Tilde{s}_{k} = [s_{k}, \boldsymbol{\phi}_{k}]$, where $s_{k}$ represents the environment (original RL) state and $\boldsymbol{\phi}_{k}$ provides the surrogate automaton representation extracted from the feature of the predictor. 
Through the augmented state $\Tilde{s}$, the problem is approximately reformulated as an MDP, such that the optimal policy is Markovian. The theoretical justification will be presented in the next section. 

Since the formal specification based reward is only available at the end of each trajectory, we employ a modified version of GRPO~\citep{guo2025deepseek} to train our actor $\pi_{\boldsymbol{\psi}}$ that incorporates the exponential weighting scheme from RWR~\citep{peters2007reinforcement}. 
In particular, given a trajectory $\tau,$ the ``raw'' advantage of the trajectory is defined to be 
\begin{align}\label{eq:rwr}
    \tilde{A}(\tau)=\exp({\beta (R(\tau))}.
\end{align}
This exponential weighting emphasizes larger rewards by assigning them disproportionately higher weights, so that under a sparse-reward landscape, the algorithm could emphasize more on positive signals, which are rare at the start.
Then, given a set of trajectories $ \mathcal{D} = \{\tau^{(n)}\}_{n=1}^N,$ GRPO normalizes the raw advantage function as follows:
\begin{align}\label{eq:advantage-normalization}
    A\left(\tau^{(n)}\right) = \frac{\tilde{A}\left(\tau^{(n)}\right)-\mathsf{mean}\left(\tilde{A}\left(\tau^{(n)}\right)\right)}{\mathsf{std}\left(\tilde{A}\left(\tau^{(n)}\right)\right)}.
\end{align}
Similarly, the GRPO loss is constructed as follows:

\begin{equation}
    \mathcal{L}_{\mathrm{GRPO}}(\boldsymbol{\psi}) = -\frac{1}{N K} \sum_{n,k} \min \bigg\{r_{n,k}(\boldsymbol{\psi}) A\left(\tau^{(n)}\right),\mathrm{clip}[r_{n,k}(\boldsymbol{\psi})]_{1-\varepsilon}^{1+\varepsilon} A\left(\tau^{(n)}\right) \bigg\},\label{eq:grpo}    
\end{equation}
where $\mathrm{clip}$ is the usual clip operator in PPO and GRPO, and $r_{n,k}$ is the importance sampling weight for the $k$-th augmented state-action pair in trajectory $\tau^{(n)}$, as follows:
\begin{align*}
    r_{n,k}(\boldsymbol{\psi}) = \frac{\pi_{\boldsymbol{\psi}}\left(\left. a_k^{(n)} \right| s_k^{(n)}, \boldsymbol{\phi}_k^{(n)} \right)}{\pi_{\boldsymbol{\psi}_{\mathrm{old}}}\left(\left. a_k^{(n)} \right| s_k^{(n)}, \boldsymbol{\phi}_k^{(n)} \right)}.
\end{align*}

\textbf{Alternating Optimization.}
As outlined in Algorithm $\ref{alg:actor_predictor}$, both the actor policy and predictor are trained on-policy. 
At each iteration $t$, the agent collects a new batch of trajectories using the current actor $\pi_{\boldsymbol{\psi}_{t}}$ and predictor $\phi_{\boldsymbol{\theta}_{k}}$. 
The two networks are trained separately with alternating optimization. 
During each iteration, we first train the actor with a frozen predictor, then train the predictor.

\section{Sufficiency of the Predictive Feature}\label{sec:theory}
In this section, we prove that the features from a well-trained predictor in our Actor-Predictor framework are a sufficient representation to learn the optimal policy under formal specifications, as a special class of Partially Observed Markov Decision Processes (POMDPs) with state space of the problem is the product space ${\cal S}_\otimes={\cal S}\times {\cal B},$
where ${\cal B}$ is the state of the automaton associated with the formal specification. 

Suppose the trajectories used to train the predictor in~\eqref{loss-bce} are from a {\em simple} policy $\mu:{\cal S}^w\times {\cal A}^{w-1}\rightarrow \mathbb{P}_{\cal A},$ which takes past $w$ step history, i.e, $(s_{k-w+1}, a_{k-w+1}, \cdots, s_{k-1}, a_{k-1}, s_k)$. 
Our results hold for all $w\in\{1, 2, \cdots, \infty\}.$ 
Given any two partial trajectories $\tau_k,$ and $\tilde{\tau}_{k'}$  truncated at step $k$ and $k'$ respectively with the same current state $s$, let $\omega$ be the mapping from the partial trajectory to the automaton state and let $b_k=\omega(\tau_k)$ and $\tilde{b}_{k'} =\omega(\tau_{k'})$, where $\tilde{b}_{k'} \neq b_k$, be the deterministic automaton state at step $k$. With a well-trained predictor and an informative policy $\mu,$ we further make the following assumptions: 

\begin{assumption}[Distinguishability] Under a well-trained predictor and informative policy $\mu,$ there exists $\epsilon>0$ such that, for any environment state $s$ and any distinct
automaton states $b$ and $\tilde b$ ($b\neq \tilde b$), the following two inequalities hold at the same time:
\begin{equation} 
\left|\sigma \left({\bm \phi}_{\bm \theta}(\tau_k)^\top {\textbf w}\right)-\mathbb{P}_\mu\left(\exists m>k,\ \omega(\tau_m)\in{\cal B}^*\mid s_k=s, b_k\right)\right|\leq \epsilon
\label{eq:pred}
\end{equation} 
\begin{equation}
\Big|\mathbb{P}_\mu\left(\exists m>k,\ \omega(\tau_m)\in{\cal B}^*\mid s_k=s, b_k\right)- \mathbb{P}_\mu\left(\exists m'>k',\ \omega(\tilde{\tau}_{m'})\in{\cal B}^*\mid \tilde{s}_{k'}=s, \tilde{b}_{k'}\right)\Big|>4\epsilon.
\end{equation} 
\label{assu:dist}  
\end{assumption}
This assumption states that, under policy $\mu,$ for any fixed environment state, distinct automaton states have distinct satisfaction probabilities. Moreover, the predictor estimates these satisfaction probabilities with an error of at most $\epsilon$. We have the following lemma under Assumptions \ref{assu:automaton} and \ref{assu:dist}, which states that the feature from the predictor is a sufficient representation of the automaton state. 
\begin{restatable}{lemma}{lempredictor}
Given trajectories $\tau_k$ and $\tilde \tau_{k'}$ such that $s_k=\tilde{s}_{k'},$ $\omega(\tau_k)=\omega(\tilde \tau_{k'})$ if and only if 
\begin{equation}
    \left|\sigma({\bm \phi}_{\bm \theta}(\tau_k)^\top{\textbf w})-\sigma({\bm \phi}_{\bm \theta}(\tilde\tau_{k'})^\top{\textbf w})\right|\leq 2\epsilon
    \label{eq:autdiff}
\end{equation} under a well-trained model $\bm \theta,$ which implies that $\sigma({\bm \phi}_{\bm \theta}(\tau_k)\not=\sigma({\bm \phi}_{\bm \theta}(\tau_k')$ if $b_k\not=\tilde{b}_{k'} $. \label{lem:predcitor}
\end{restatable}

The proof of this lemma is included in the supplementary material. From this lemma, we can establish the following theorem. 
\begin{restatable}{theorem}{maintheorem}
There exists a Markov policy $\tilde \pi^*: {\cal S}\times {\Phi}_{\bm \theta}\rightarrow \mathbb{P}_{\cal A}$ such that 
$$\mathbb E_{\tilde \pi^*}\left[\sum_k r(s_k, b_k)\right]=\max_{\pi\in \Pi}\mathbb E_{\pi}\left[\sum_k r(s_k, b_k)\right],$$ where $\Pi$ is the set of all history-dependent policies. \label{thm:policy}
\end{restatable}
\begin{proof}
Note that the problem is a fully observed MDP with $(s_k, b_k).$ Therefore, there exists a stationary Markov policy $\pi^*: {\cal S}_\otimes\rightarrow \mathbb P_{\cal A}$ such that  
$$\mathbb E_{\pi^*}\left[\sum_k r(s_k, b_k)\right]=\max_{\pi\in \Pi}\mathbb E_{\pi}\left[\sum_k r(s_k, b_k)\right].$$

Now given $\pi^*,$ we define $\tilde \pi^*$ such that at step $k,$
$\tilde\pi^*(s_k,{\bm \phi}_k)=\pi^*(s_k,b)$ where $b$ is the automaton state such that
\begin{align}
\left|\sigma \left({\bm \phi}_k^\top{\textbf w}\right) - \mathbb{P}_\mu\left(\exists K\geq m>k,\ \omega(\tau_m)\in{\cal B}^*\mid s_k=s, b_k\right)\right|\leq \epsilon.
\end{align} Now according to the previous lemma, if $\bm \phi_k={\bm \phi}_{\bm \theta}(\tau_k),$ then $b=b_k,$ the correct automaton state. 
\end{proof}

This theorem states that with a well-trained predictive model, we can find a Markov policy on the state space ${\Phi}_{\bm \theta}\times {\cal S}$ which achieves the maximum reward for the original problem on $\cal S_\otimes$, which justifies our framework. 
Unlike classical PSRs, which seek a predictive representation sufficient for all future observations, our representation only needs to be sufficient for predicting task satisfaction and therefore may be substantially lower-dimensional.
 
\section{Experiments}
\label{sec:experiments}

\begin{wrapfigure}{r}{0.43\columnwidth}
    \vspace{-8pt}
    \centering
    \includegraphics[width=\linewidth]{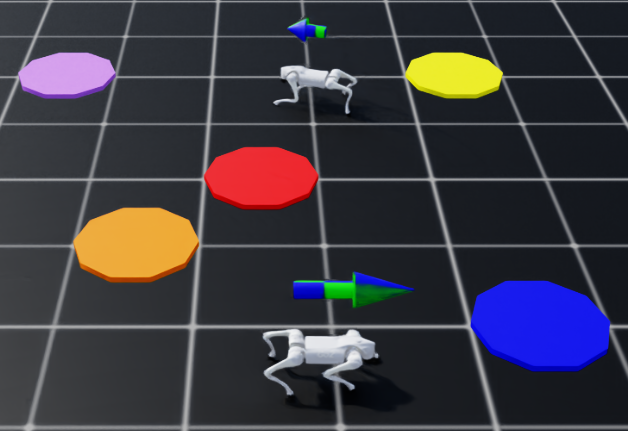}
    \vspace{-5pt}
    \caption{The quadruped environment.}
    \label{fig:isaac}
    \vspace{-6pt}
\end{wrapfigure}

We evaluate our Actor-Predictor architecture across four key dimensions: (1) performance against baselines with explicit automaton access, (2) expressivity on non-regular formal specifications, (3) feature transferability across specifications, and (4) latent state alignment relative to ground truth automaton states.

\textbf{Quadruped.} Implemented in Isaac Lab~\citep{mittal2025isaaclab}, this environment uses a Unitree quadruped.
Control is hierarchical: a high-level policy outputs a discrete target velocity (linear $\pm 1, m/s$ or heading ($\pm \pi/2, rad/s$) mapped to joint torques by a frozen low-level controller. 
In tasks involving the collision predicate $C$, a second robot patrols a circle of radius $\sqrt{2}$ m around the red landmark (see Figure \ref{fig:isaac}). 
The observation vector $\mathbf{s}$ stacks projected gravity $\mathbf{g}$, local 2D landmark positions (including $\mathbf{o}$ only when present in logic), and robot $\mathbf{r}_2$ (zero-padded when absent).
The temporal logic specifications are
$\varphi_{1}=\mathbf{G}(\mathbf{F}(B))\land \mathbf{G}(\mathbf{F}(P))\land \mathbf{G}(O \rightarrow \mathbf{F}(Y)) \land \mathbf{G}(\neg R) \land \mathbf{G}(\neg C)$, 
$\varphi_{2}=\mathbf{G}\big(\mathbf{F}(P \land \mathbf{X}(\mathbf{F}(Y \land
\mathbf{X}(\mathbf{F}(B)))))\big) \land \mathbf{G}(\neg R)$, and 
$\varphi_{3}=\mathbf{G}(\mathbf{F}(B))\land \mathbf{G}(\mathbf{F}(P))\land \mathbf{G}(\neg R) \land \mathbf{G}(\neg C)$
where $B$, $P$, $O$, $Y$, and $R$ denote the entry of the blue, purple, orange, yellow, and red landmarks, respectively, and $C$ is a collision $\|\mathrm{pos_{agent}}-\mathrm{pos_{robot2}}\|_{2} \leq 0.75$ m.
We set batch size $512$, decision frequency 5\, Hz, and episode lengths of 135\, s for $\varphi_{1}, \varphi_{3}$ and 150\, s for $\varphi_{2}$.
The agent receives a reward of $+1$ each time the specification is newly satisfied. The APs reset after visiting an accepting state. 

\textbf{Grid World.} A $10 \times 10$ discrete grid world where each episode begins at $(0,0)$. 
The agent takes discrete cardinal actions.
Four landmarks are placed at blue $(9,0)$, purple $(0,9)$, red $(4,4)$, and yellow $(9,9)$. 
This grid world environment was used to evaluate the non-regular formal specification $\mathcal{L}=\{a^{n}b^{n}\}$. 
This is a foundational example in formal language theory that cannot be represented with a finite automaton. Tasks $a$ and $b$ are two subtasks, each requiring the agent to reach a target location on the grid (Blue and Purple respectively) and then return to a specified goal location (Yellow). 
The red region must be avoided.
These tasks must be completed an {equal number of times eventually, and task $a$ must never be completed after task $b$} (e.g., $ab,$ $aabb$). 
We train with batch size 4096 over $400$-step episodes.

\textbf{Baselines.} We compare our approach against four standard baselines:
\textbf{LCER}~\citep{voloshin2023eventual} uses an MLP policy taking state and automaton inputs, reusing transitions across automaton states via counterfactual relabeling with eventual discounting. 
\textbf{CyclER}~\citep{shah2024ltl} shapes rewards by maximum path completion using cycle experience replay and eventual discounting. 
\textbf{PPO}~\citep{schulman2017proximal} serves as a standard actor-critic baseline using the same Mamba architecture for fair comparison. 
\textbf{GRPO}~\citep{shao2024deepseekmath} is a critic-free algorithm estimating advantages via trajectory-group reward normalization under the Mamba architecture.

\begin{figure}[t]
    \vspace{-10pt}
    \centering

    \begin{minipage}[b]{0.315\linewidth}
        \centering
        \includegraphics[width=\linewidth]{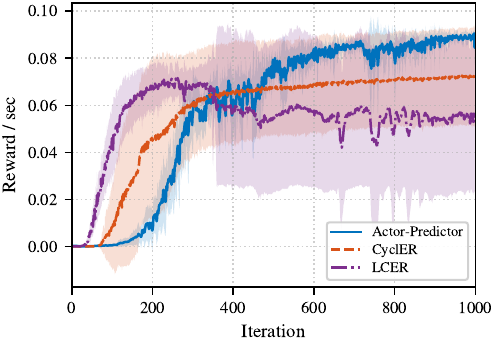}
        \vspace{-2mm}

        {\small (a)}
    \end{minipage}%
    \hfill
    \begin{minipage}[b]{0.315\linewidth}
        \centering
        \includegraphics[width=\linewidth]{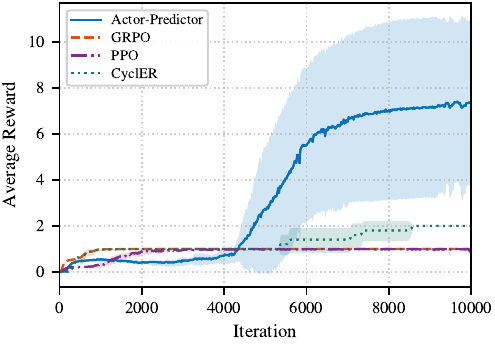}
        \vspace{-2mm}

        {\small (b)}
    \end{minipage}%
    \hfill
    \begin{minipage}[b]{0.315\linewidth}
        \centering
        \includegraphics[width=\linewidth]{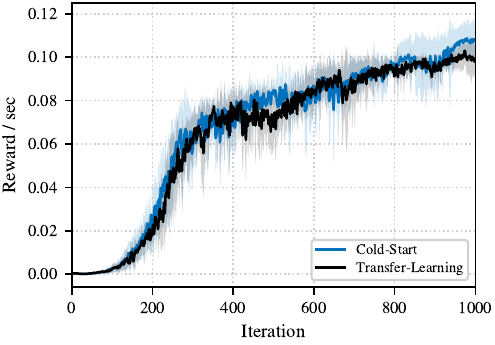}
        \vspace{-2mm}

        {\small (c)}
    \end{minipage}

    \caption{(a) Comparison of Actor-Predictor, CyclER, and LCER on $\varphi_{1}$ ($x$-axis is number of gradient updates). (b) Comparison of Actor-Predictor, GRPO, PPO, and CyclER on $\mathcal{L} = a^{n}b^{n}$. (c) Transfer learning from $\varphi_{2}$ to $\varphi_{3}$, where the actor is trained in both Cold-Start and Transfer-Learning, while the predictor is trained only in Cold-Start.}
    \label{fig:main_results}
    \vspace{-10pt}
\end{figure}
\subsection{Performance Comparison with the Baselines}

We compare single-head Actor-Predictor against CyclER and LCER under specification $\varphi_{1}$ in the quadruped environment (Fig.~{\ref*{fig:main_results}(a)}).
Actor-Predictor achieves the highest average reward ($0.0905 \pm 0.0027$), outperforming CyclER ($0.0724\pm0.0216$) and LCER ($0.0566\pm0.0328$).
Actor-Predictor has the slowest convergence because, unlike the baselines, it is not explicitly provided with the automaton state and there is no reward shaping, making initial rewards much more sparse.

Next, we use the grid world with the non-regular specification $\mathcal{L} = \{a^{n}b^{n}\}$. 
Since $\mathcal{L}$ is not regular, CycLER uses a finite automaton that creates an upper bound of $n$ to $15$ (Fig.~{\ref*{fig:main_results}(b)}).
Even with this truncation at $N=15$, the automaton has 137 states, which shows as the maximum count $n$ suffer from state space explosion. 
Our Actor-Predictor architecture bypasses explicit automaton construction entirely, avoiding this scalability limitation. 
Actor-Predictor substantially outperforms all baselines, achieving a final average reward of $7.37 \pm 3.56$: four of five seeds reach $n=9$, while the
remaining seed converges to $ab$. CyclER reaches $2.0 \pm 0.0$, and GRPO and
PPO reach $1.0 \pm 0.0$ and $0.98 \pm 0.02$ respectively, corresponding to the
same $ab$ optimum.

The next experiment tests whether predictor features transfer across
specifications (Fig.~{\ref*{fig:main_results}(c)}). The ``Transfer-Learning'' Actor-Predictor was first trained on $\varphi_{2}$ with
two prediction heads for 1,000 iterations, after which the predictor was frozen
and the actor was randomly re-initialized. The actor was then trained on
$\varphi_{3}$. 
The ``Cold-Start'' agent was trained on $\varphi_{3}$ from scratch
with one prediction head. 
The positions of $R$, $B$, and $P$ were shared
between the two specifications; $\varphi_{3}$ additionally involves a second
robot, whereas $\varphi_{2}$ does not.
Cold-Start reaches a final reward of $0.1080 \pm 0.0094$, while Transfer-Learning reaches $0.0980 \pm 0.0078$. 
This indicates the transferability of the learned features across different specifications. 
By freezing the Mamba predictor, we also achieve a ~3.48 GB reduction in peak VRAM overhead per step. 
Empirically, while an LSTM predictor exhibits similar memory savings, the Transformer predictor yields more than twice the peak VRAM savings.

\subsection{Automaton State Recovery From Hidden State}
We train a linear classifier on 2,048 trajectories to recover the automaton state $b_{t}$ from the predictor hidden state $h_{t}$, using an 80/20 train-test split. The high-recover accuracy in Table~\ref{tab:state_recovery} indicates that $h_{t}$ captures the task dynamics, empirically supporting Lemma~\ref{lem:predcitor}.

\begin{table}[ht]
    \centering
    \vspace{-10pt}
    \caption{\small Automaton state recovery accuracy across environments (\%).}
    \label{tab:state_recovery}
    \small
    \setlength{\tabcolsep}{8pt}
    \begin{tabular}{ccc}
    \toprule
    \textbf{Quadruped ($\varphi_{1}$)} & \textbf{Quadruped ($\varphi_{2}$)} & \textbf{Grid World ($\mathcal{L}=\{a^{n}b^{n}\}$)} \\
    \midrule
    $99.63 \pm 0.14\%$ & $99.55 \pm 0.22\%$ & $94.12 \pm 10.30\%$ \\
    \bottomrule
    \end{tabular}
    \vspace{-10pt}
\end{table}

\begin{figure}[t]
    \vspace{-10pt}
    \centering
    \setcounter{enumi}{0}

    \begin{minipage}[t]{0.235\linewidth}
        \centering
        \includegraphics[width=\linewidth]{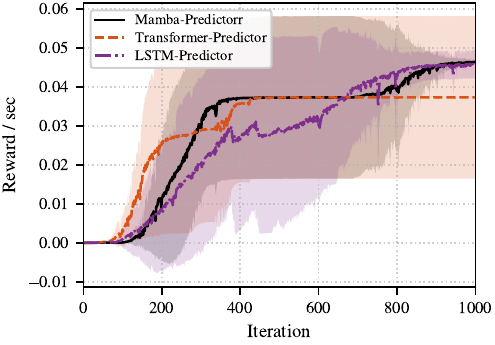}
        \refstepcounter{enumi}
        \label{subfig:task1}
        \smallskip

        (\alph{enumi}) Ablation over predictor architectures on $\varphi_{2}$
    \end{minipage}%
    \hfill
    \begin{minipage}[t]{0.235\linewidth}
        \centering
        \includegraphics[width=\linewidth]{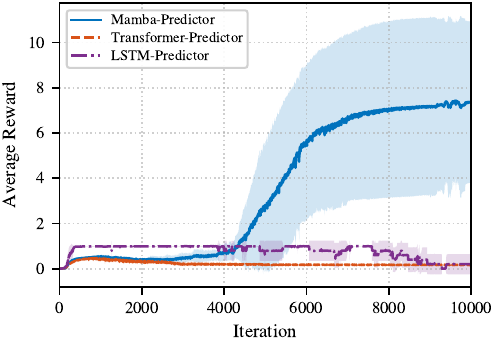}
        \refstepcounter{enumi}
        \label{subfig:task2}
        \smallskip

        (\alph{enumi}) Ablation over predictor architectures on $\mathcal{L}=\{a^{n}b^{n}\}$
    \end{minipage}%
    \hfill
    \begin{minipage}[t]{0.235\linewidth}
        \centering
        \includegraphics[width=\linewidth]{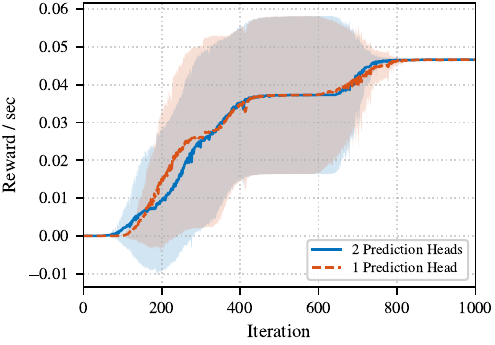}
        \refstepcounter{enumi}
        \label{subfig:task3}
        \smallskip

        (\alph{enumi}) Ablation over the number of prediction heads on $\varphi_{2}$
    \end{minipage}%
    \hfill
    \begin{minipage}[t]{0.235\linewidth}
        \centering
        \includegraphics[width=\linewidth]{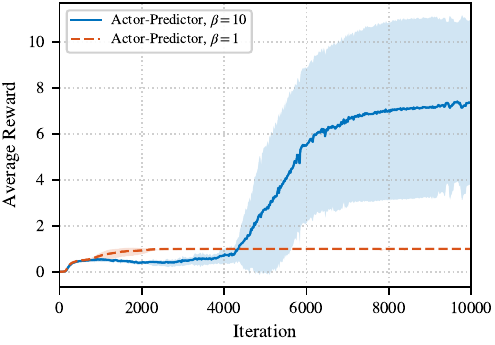}
        \refstepcounter{enumi}
        \label{subfig:task4}
        \smallskip

        (\alph{enumi}) Actor-Predictor with different $\beta$ values on $\mathcal{L}=\{a^{n}b^{n}\}$
    \end{minipage}

    \caption{Ablation over the Actor-Predictor Design Choices}
    \label{fig:ablation_results}
    \vspace{-10pt}
\end{figure}
\subsection{Ablation}

We ablate key design choices in our method: the number of prediction heads, the predictor architecture, and exponential weighting. 
We first compared Mamba, Transformer, and Long Short-Term Memory (LSTM) predictors, each paired with an MLP actor, on the grid world using the non-regular language $\mathcal{L}=\{a^{n}b^{n}\}$ (Fig.~{\ref*{fig:ablation_results}(a)}) and on the quadruped environment using the regular specification $\varphi_{2}$ (Fig.~{\ref*{fig:ablation_results}(b)}). 
Under regular constraints, Mamba, Transformer, and LSTM achieve comparable rewards ($0.0464 \pm 0.0005$, $0.0373 \pm 0.0209$, $0.0458 \pm 0.0038$). Under non-regular constraints, Mamba reaches $7.37 \pm 3.56$, outperforming Transformer ($0.17 \pm 0.02$) and LSTM ($0.20 \pm 0.44$).
Mamba has the best performance for both non-regular and regular constraints.

We also ablate the number of prediction heads: each head predicts the
satisfaction of one conjunct.
For $\varphi_{2}$, we predict the full specification along with
$\varphi_{2,1}=\mathbf{G}\big(\mathbf{F}(P \land \mathbf{X}(\mathbf{F}(Y \land
\mathbf{X}(\mathbf{F}(B)))))\big)$ and $\varphi_{2,2} = \mathbf{G}(\neg R)$.
The two-headed ($0.0466\pm0.000$) and single-head ($0.0466\pm0.001$) variant have nearly the same reward in Fig.~{\ref*{fig:ablation_results}(c)} shows the compositionality of the specification.

The final ablation investigates the effect of exponential weighting in the
$a^{n}b^{n}$ task. 
The unweighted variant ($\beta=1$) receives an average reward of $1.0 \pm 0.0$,
converging to the $ab$ local optimum, while the exponentially weighted variant
($\beta=10$) reaches $7.4 \pm 3.6$ (Fig.~{\ref*{fig:ablation_results}(d)}). 
The probability of increasing $n$ is more challenging than in the previous step, necessitating exponential weighting and encouraging exploration to higher $n$. 

\section{Conclusion}
RL with formal specification is challenging because of the sparse reward nature.
We propose Actor-Predictor, a framework that learns a surrogate representation of logic progress without knowledge or observability of the automaton. 
We show theoretically that, with sufficient predictive accuracy, the learned representation is equivalent to the automaton state. 
Empirically, Actor-Predictor outperforms automaton-based baselines on LTL specifications without access to the automaton, learns non-regular specifications such as $\mathcal{L}=\{a^{n}b^{n}\}$, and learns environmental features that can be transferred across specifications.

\section*{AI Disclosure}
LLMs were used for coding and proofreading assistance. The authors reviewed all outputs and take full responsibility for the content of this work.

\bibliographystyle{plainnat}
\bibliography{references,inlab-refs}

\newpage
\clearpage

\appendix

\section{Proof of Lemma \ref{lem:predcitor}}
Given $s_k=\tilde{s}_{k'}=s$ and $\omega(\tau_k)=\omega(\tilde{\tau}_{k'})=b,$ 
the success probabilities of $\tau$ and $\tilde \tau$ are the same because they have the same probability distribution over the automaton states at termination, so the success probabilities are the same, or 

\begin{align*}
\mathbb{P}_\mu\left(\exists m>k,\ \omega(\tau_m)\in{\cal B}^*\mid s_k=s, b_k\right)=\mathbb{P}_\mu\left(\exists m'>k',\ \omega(\tilde{\tau}_{m'})\in{\cal B}^*\mid \tilde{s}_{k'}=s, \tilde{b}_{k'}\right)=p,
\end{align*}
which implies that 
$$|\sigma \left({\bm \phi}_{\bm \theta}(\tau_k)^\top{\textbf w}\right)-p|\leq \epsilon\quad \hbox{and}\quad |\sigma \left({\bm \phi}_{\bm \theta}(\tilde\tau_{k'})^\top{\textbf w}\right)-p|\leq \epsilon,$$
\hbox{by the triangle inequality} $$\left|\sigma({\bm \phi}_{\bm \theta}(\tau_k)^\top{\textbf w})-\sigma({\bm \phi}_{\bm \theta}(\tilde\tau_{k'})^\top{\textbf w})\right|\leq 2\epsilon.$$

\noindent Now given $s_k=\tilde s_{k'}=s$ 
with a well-trained model, we have
\begin{align}
\left|\sigma \left({\bm \phi}^\top_{\bm \theta}(\tau_k){\textbf w}\right) - \mathbb{P}_\mu\left(\exists m>k,\ \omega(\tau_m)\in{\cal B}^*\mid s_k=s, b_k\right)\right|\leq \epsilon
\end{align} and
\begin{align}
\left|\sigma \left({\bm \phi}^\top_{\bm \theta}(\tilde\tau_{k'}){\textbf w}\right) - \mathbb{P}_\mu\left(\exists m'>k',\ \omega(\tilde{\tau}_{m'})\in{\cal B}^*\mid \tilde{s}_{k'}=s, \tilde{b}_{k'}\right)\right|\leq \epsilon,
\end{align} applying the triangle inequality again
\begin{align}
&\Big|\mathbb{P}_\mu\left(\exists m>k,\ \omega(\tau_m)\in{\cal B}^*\mid s_k=s, b_k\right)-\mathbb{P}_\mu\left(\exists m'>k',\ \omega(\tilde{\tau}_{m'})\in{\cal B}^*\mid \tilde{s}_{k'}=s, \tilde{b}_{k'}\right)\Big|\leq 4 \epsilon,
\end{align}  which implies that $\omega(\tau_{k})=\omega(\tilde \tau_{k'}).$

\section{Experimental Details}

\subsection{Hardware}
All experiments were performed on a single NVIDIA RTX PRO 6000 Blackwell.

\subsection{Details}
We began from standard Proximal Policy Optimization (PPO) and Adam defaults, together with the values reported in the Cycler Experience Replay (CyclER) paper, and adjusted them to perform well across all algorithms. 
All experiments in both the grid world and Quadruped environments were run with random seeds 0–4. 
One-step temporal-difference (TD) returns were used to estimate the advantage for vanilla PPO and PPO with Eventual Discounting. 
PPO with Eventual Discounting was used with the reward shaping of CyclER and LTL-guided Counterfactual Experience Replay (LCER).

\paragraph{Grid World}

The grid world had a batch size of 4096 environments, the maximum that fit on a single NVIDIA RTX PRO 6000 Blackwell.

We tested time horizons of \{350, 400, 500, 600, 700\} and found 400 to be the shortest horizon that consistently scaled with $n$.
The actor learning rate was swept over \{2e-5, 4e-5, 6e-5, 8e-5, 1e-4, 3e-4, 5e-4\}.
The entropy bonus was swept over \{1e-5, 1e-4, 1e-3, 5e-3, 1e-2\}.
The clip ratio was tested over \{0.1, 0.2, 0.3\}, and PPO epochs over \{2, 3\}.
The discount factors tested were 0.99 and 0.999.
$\beta$ was tested at \{1,2,4,6,7,8,10,12\}, with $\beta=10$ giving the best performance. 
We used a Lagrangian $\lambda$ of 10, guided by the original CyclER paper.
Final values for each algorithm are reported in Table \ref{tab:gridparam}.

\begin{table}[H]
\centering
\begin{tabular}{lcccc}
\toprule
Parameter & AP & CyclER & GRPO & PPO \\
\midrule
Actor lr       & 4e-5 & 4e-5 & 4e-5 & 4e-5 \\
Critic lr  & -- & 1e-3 & -- & 1e-3 \\
Predictor lr  & 1e-3 & -- & -- & -- \\
Entropy bonus $\alpha$ & 5e-3 & 5e-3 & 5e-3 & 5e-3 \\
Clip ratio $\epsilon$ & 0.2 & 0.2 & 0.2 & 0.2  \\
PPO epochs         & 3 & 3 & 3 & 3 \\
Discount $\gamma$  &  -- & 0.99 & -- & 0.99  \\
$\beta$         & 10 & -- & -- & --   \\
Max grad. norm  & None & 0.5 & None & None       \\
Lagrangian $\lambda$ & -- & 10 & -- & -- \\
\bottomrule
\end{tabular}
\caption{Hyperparameters for each algorithm in the grid world environment. 
All experiments use the grid world configuration described in the Experiments section.}
\label{tab:gridparam}
\end{table}

CyclER uses an MLP actor and critic, each with three hidden layers of size 128, 64, and 64.
We evaluated three predictor architectures for the actor: Mamba, Transformer, and Long Short-Term Memory (LSTM).
The Mamba predictor uses a local temporal convolution of width 4, a state dimension of 8, and an internal expansion factor of 1. 
The Transformer predictor uses a 4-head transformer with Rotary Positional Embeddings (RoPE), 4-dimensional query and value projections, and a feed-forward layer expanded by a factor of 4.
The LSTM predictor is single-layer with a hidden state size of 416 and an intermediate convolution of width 4.
The actor for all AP (Actor-Predictor) methods uses two hidden layers of sizes 128 and 64. 
GRPO and PPO both use a Mamba actor with a state dimension of 8, a convolution of width 4, and an expansion factor of 1.
PPO uses a Mamba-based critic with the same configuration as the actor.

\paragraph{Quadruped}

In the Quadruped environment, parameters not explicitly swept below were adopted from the grid world.

The control frequency was 5 Hz.
The time horizons for $\varphi_{1}$ and $\varphi_{3}$ were both 135 s (675 steps), while the time horizon for $\varphi_{2}$ was 150 s (750 steps).
We found that shorter time horizons made the reward too sparse.
We set the batch size to 512, which allowed 1000 training iterations to complete in reasonable time on a single NVIDIA RTX PRO 6000 Blackwell. 
We selected the actor learning rate by starting at 1e-4 and increasing in steps of 1e-4, choosing the highest value that maintained stable training (3e-4).
We swept the entropy bonus over \{1e-5, 1e-4, 1e-3, 1e-2, 5e-2\}, PPO epochs over \{1, 2, 3\}, and the clip ratio $\epsilon$ over \{0.1, 0.2, 0.3\}, selecting the values that gave the most stable training. 
No Lagrangian $\lambda$ is needed for CyclER, as there are no MDP rewards. 
Final values for each algorithm are reported in Table \ref{tab:IsaacEnv}.

\begin{table}[H]
\centering
\begin{tabular}{lccc}
\toprule
Parameter & AP & CyclER & LCER \\
\midrule
Actor lr       & 3e-4 & 3e-4 & 3e-4 \\
Critic lr       & -- & 1e-3 & 1e-3 \\
Predictor lr       & 1e-3 & -- & -- \\
Entropy bonus $\alpha$    & 1e-4 & 1e-4 & 1e-4 \\
Clip ratio $\epsilon$     & 0.2  & 0.2  & 0.2  \\
PPO epochs         & 3    & 3    & 3    \\
Discount $\gamma$         & -- & 0.99 & 0.99 \\
$\beta$                   & 10   & --   & --   \\
Max grad. norm         & None & 0.5  & 0.5  \\
\bottomrule
\end{tabular}
\caption{Hyperparameters for each algorithm in the Quadruped environment. All experiments use the Quadruped task configurations described in the Experiments Section.}
\label{tab:IsaacEnv}
\end{table}

The CyclER and LCER methods use an MLP actor and critic, each with three hidden layers of size 64. 
AP uses a single Mamba layer to make predictions.
The Mamba predictor uses a local temporal convolution of width 4, a state dimension of 8, and an internal expansion factor of 1.
The actor has two hidden layers, each of size 64.

\end{document}